# Toward Experiential Utility Elicitation for Interface Customization

**Bowen Hui**
Department of Computer Science
University of Toronto
Toronto, Ontario, Canada

**Craig Boutilier**
Department of Computer Science
University of Toronto
Toronto, Ontario, Canada

## Abstract

User preferences for automated assistance often vary widely, depending on the situation, and quality or presentation of help. Developing effective models to learn individual preferences online requires domain models that associate observations of user behavior with their utility functions, which in turn can be constructed using utility elicitation techniques. However, most elicitation methods ask for users' *predicted* utilities based on hypothetical scenarios rather than more realistic *experienced* utilities. This is especially true in interface customization, where users are asked to assess novel interface designs. We propose experiential utility elicitation methods for customization and compare these to predictive methods. As experienced utilities have been argued to better reflect true preferences in behavioral decision making, the purpose here is to investigate accurate and efficient procedures that are suitable for software domains. Unlike conventional elicitation, our results indicate that an experiential approach helps people understand stochastic outcomes, as well as better appreciate the sequential utility of intelligent assistance.

## 1 Introduction

Intelligent software customization has become increasingly important as users are faced with larger, more complex applications. For a variety of reasons, software must be tailored to specific individuals and circumstances [21]. For example, adaptive interfaces are critical as different users may: require different functionality from multi-purpose software [5]; prefer different modes of interaction; use software on a variety of hardware devices [12]; or, due to expanding software complexity, require online and automated help to identify and master different software functions [16]. In the latter case, a system should ideally adapt the help it provides and the decision to interrupt a user [15] to account for specific user preferences.

In this paper, we focus on *interface customization* where the attributes of interface widgets (e.g., location, transparency, and functionality) are automatically tailored to the needs of specific users. In particular, we are interested in intelligent systems that learn to predict user goals over time based on observed user behavior, and suggest ways (e.g., through interface customization) to help the user complete the desired goal. Considerable work has been devoted to the prediction of user needs and goals (e.g., [16; 1; 28; 3] among others), much of it is focused on developing probabilistic models of user goals. Less emphasis has been placed on assessing user preferences for software interaction and customization (for exceptions, see [11; 17]). However, accounting for user preferences is critical to good interface customization. For instance, consider automated word completion [10]. Some users may prefer single-word suggestions, while others may prefer several different suggestions. Similarly, some users may be satisfied with "partial help" (e.g., a partially correct word that saves a few keystrokes) while others may wish to use only completely correct completions. These preferences, and more importantly, a user's *strength of preference*, are needed to make suggestion decisions: preferences must be weighed against the probability of specific user goals.

In this paper, we evaluate the effectiveness of a variety of preference elicitation techniques for interface customization. While elicitation may not be used directly (i.e., online) during application use, most adaptive systems will make indirect assessments of user preferences (e.g., [32; 30; 8; 19]). However, even indirect assessment methods require some knowledge of the range of possible user preferences and how those are (perhaps stochastically) related to observable behavior. In these cases, offline preference elicitation for different customizations can provide valuable data for the design of an online system.

Most existing literature on preference elicitation in AI, as well as the majority of that in behavioral decision theory, assumes that people "know" their preferences *a pri-*

*ori*. However, often users have not encountered, nor even considered, the hypothetical situations typically posed in the elicitation process. This is especially true of software customization, since the situations involve novel interfaces. Under these circumstances, people may report their *predicted utilities* by conceptualizing the posed scenarios and forecasting their own preferences. However, what people "think they like" can systematically differ from what they "actually like" [23]. For example, someone who has not actually engaged in a system that offers a range of partial word completion suggestions may predict that they dislike the interface in a particular circumstance, but in fact like it when they experience it (or vice versa).

For this reason, we propose a novel *experiential elicitation* approach for interface customization. Our elicitation "queries" allow users to assess *experienced utilities* [25] by providing simple, hands-on tasks and system suggestions or customizations, drawn from a particular distribution. We adapt standard elicitation approaches to incorporate such *experiential queries*. Our approach also overcomes some of the difficulties with well-established procedures (e.g., standard gambles) that involve probabilities over a distribution of outcomes. We explore this new approach in the context of a specific customization task—the suggestion of highlighting options in PowerPoint — and show that experiential elicitation offers a more accurate means of assessing quantitative tradeoffs in preferences. Unfortunately, one drawback of experiential queries is the time they take. To counteract this, we also propose two hybrid models that attempt to assess experienced utilities somewhat more (time) efficiently. Our results show that one hybrid procedure provides a good approximation to the experiential approach in a much more effective manner.

In Section 2, we outline the basic customization domain, and describe the underlying decision-theoretic model used to provide assistance to users. We describe essential background on preference elicitation in Section 3. Our approach to experiential elicitation for interface customization is presented in Section 4, as is our empirical evaluation. In Section 5 we consider two hybrid approaches that accelerate the experiential process, *primed* and *primed+* elicitation, and evaluate their effectiveness. The key benefit in our experiential approach is that it enables users to better interpret queries and assess the sequential utility of an interface via hands-on experience. As a result, our approach can provide a more accurate picture of the user's preferences.

## 2 The Customization Domain

We are interested in developing intelligent systems that perform online interface customization based on user preferences and user needs. We focus on a concrete form of customization as an example to illustrate our approach, but note that the general principles apply more broadly. We use this domain and application to motivate the need for preference elicitation for interface customization and to highlight the difficulties with standard elicitation methods. Our main contributions are discussed in Sections 4 and 5, but will apply more broadly than the task considered here.

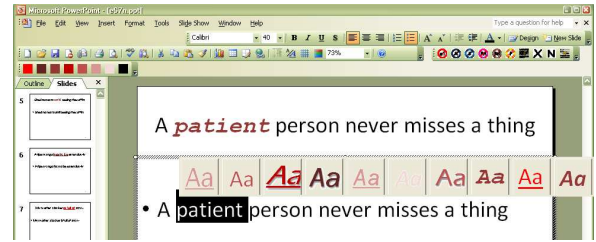

Figure 1: Icon suggestions to help the user in a highlighting task implemented as part of PowerPoint 2003.

### 2.1 The Highlighting Task

Consider a user authoring slides in PowerPoint who wishes to highlight important phrases and new terminology by applying a particular font stylization. For aesthetic reasons, the user tends to choose from just a few highlighting styles that are consistent throughout a presentation, and possibly across multiple presentations. This consistency provides the opportunity for an intelligent system to observe and learn user-specific patterns, so that it can offer useful suggestions in the future. Figure 1 shows an example toolbar with 10 icons, each suggesting a particular set of font characteristics that can be applied to highlight the selected phrase "patient." Many repetitive tasks in PowerPoint and other software can benefit from automated suggestions designed to minimize user effort.

We assume that in a highlighting task, the user has a certain style in mind, which involves changing some number of font *features* to make it stand out from the rest of the text. We define the *complexity* of a highlighting style as the number of font feature values required to create it. In other words, this is the number of events that the user must execute to make a phrase different from neighboring text or some baseline default font style. For example, relative to "plain" black text, a bold, italicized font has complexity 2.

To assist the user with the highlighting task, our system can suggest a toolbar with a number of icons, each offering some combination of font features shown by the characters "Aa" (as illustrated in Figure 1). The user may: select one of these icons to apply the associated font style; complete the highlighting task purely manually using the mouse (e.g., click on the Bold icon) or shortcut keys (e.g., press Ctrl-B); or accept a suggestion, but further refine it manually by applying (or undoing) additional font characteristics. If the toolbar is ignored (e.g., the user continues typing), it disappears after a short time.

## 2.2 Predictive Model & Assistance Decisions

Intuitively, the value of highlighting suggestion depends on the amount of savings it offers relative to manual task completion. There are also costs to suggestions: interruption, processing costs, mode switching, etc. We discuss the relative value and costs of a set of toolbar suggestions below. Notice, however, that the value of a suggestion cannot be known with certainty: the system can only make a stochastic prediction about the user's true intended goal. These predictions must be weighed against the overall costs and benefits of making (or not making) a suggestion.

While our aim is not to discuss predictive models, we give a sketch of our system model in order to place our elicitation results in the appropriate context. We focus specifically on *highlighting goals* in which the user desires a certain style (combination of font attributes). The system observes past user events and learns user-specific styles, which are stored in a goal library. For each goal, the system creates a probabilistic event model (a stochastic automaton), and at runtime maintains a distribution over goals given the stream of observed user events. Suggestions are made (or not) decision-theoretically using a partially observable Markov decision process (POMDP) [2] to tradeoff goal probabilities with the costs and benefits of various suggestions [19] given specific user preferences. We elaborate on the cost model in Section 2.3, as well as how user preferences can be incorporated into a POMDP in Section 3.

## 2.3 The Value and Costs of Suggestions

By selecting one of the suggested icons, the user saves the effort of manually completing the task herself (or some part of the task). *Savings* is an objective measure of help quality, reflecting the number of steps/actions a user avoids by accepting a suggestion. Common examples that provide savings are auto-completion and the Office 2007 minitoolbar. Research in human-computer interaction (HCI) suggests that the manual effort of interaction (e.g., moving the mouse, typing, mode switching) is a function of the user's actions, the number of such actions, and the modes used to execute them [6]. The quality of help actions in adaptive systems can be defined similarly, capturing the difference between manual effort required by the user with and without help. We define the quality $Q(i|g)$ of a suggestion icon $i$ (given a goal $g$) to be this difference. Since we expect users to pick the best icon available, we define the quality of toolbar $t$ to be $Q(t|g) = \max_{i \in t} Q(i|g)$, the maximum effort saved by any icon in the toolbar.[1]

The subjective value of help may vary depending on certain user features, such as neediness and distractibility [16; 19], frustration/distress [26; 9; 19], and independence [19].

---
[1]Of course, since the goal is only known stochastically, expected quality must be computed relative to the system's beliefs about the user's current highlighting goal.

In other words, quality of the suggestion alone may not predict system utility. For example, someone who currently needs help with a difficult task may benefit greatly from partial suggestions that help the user identify the next steps, while someone who is highly independent may not even accept (or find value in) suggestions of perfect quality. In other words, the *perceived utility* of automated help is a function of certain user characteristics, such as how much help the user needs or how independent the user is. In a probabilistic model, these characteristics are hidden user variables that need to be estimated online based on sequences of observed user behavior. In our initial design, we focus on neediness only. Examples of observable characteristics that a system can use to infer the user's neediness level include the user being stuck (i.e., pausing during task activity) or looking for help (i.e., browsing without selection).[2] These observations often arise in difficult tasks. Therefore, we define the user's level of neediness $N(g)$ as a function of how difficult the goal $g$ is to the user. (We explain how we simulate user neediness in a controlled experiment in Section 4.2.)

Apart from potential savings, toolbar suggestions also have associated costs. In general, *information processing* refers to the user scanning and evaluating a set of items of similar nature. Common interface examples that require information processing are menus and toolbars. Research in HCI suggests that the time it takes a user to process a set of items in an interface is a function of the number of items and the search strategy used based on the expertise level of the user [13; 22; 14]. In adaptive toolbars, the icons are always changing, so users cannot develop a search strategy to minimize processing time. To model processing time, we focus only on the number of items in an adaptive toolbar. We define length $L(t)$ to be the number of icons in $t$.

Depending on the specific system action, costs other than information processing may be relevant in determining system utility. We refer interested readers to [20] for a detailed discussion of interaction cost models.

## 3 Preference Elicitation

In order for the system to choose a good toolbar to help the user, it needs a model of the user's preferences for possible suggestions. The value of a suggestion $t$ depends on the user's utility function with respect to user neediness $N$, toolbar length $L$, and suggestion quality $Q$. Let $O$ be the set of possible outcomes over the values of these three attributes. We use notation such as $n1, l5, q4$ to represent the outcome with the user's neediness level at $1$ (high) and the toolbar showing five icons with toolbar quality $4$. The need for a utility function (rather than qualitative preferences) should be apparent given the stochastic nature of goal estimation.

---
[2]Bayesian models exist for learning neediness [16; 19].

A user's preferences for particular outcomes, including their strength of preference, can be represented by a *utility function*, $u : O \to \mathbb{R}$, where $u(o_i) > u(o_j)$ iff $o_i$ is preferred to $o_j$, and $u(o_i) = u(o_j)$ iff the user is indifferent between $o_i$ and $o_j$. For convenience, we normalize utilities to the interval $[0, 1]$, defining $o^\top$ to be the best outcome with $u(o^\top) = 1$ and $o_\bot$ to be the worst outcome with $u(o_\bot) = 0$. A utility function can be viewed as reflecting qualitative preferences over *lotteries* (distributions over outcomes) [27], with one lottery preferred to another iff its expected utility is greater. Let $SG(p) = \langle p, o^\top; 1\text{-}p, o_\bot \rangle$ denote a *standard gamble*, a specific (parameterized) lottery where $o^\top$ is realized with probability $p$ and $o_\bot$ is realized with probability $1 - p$. The expected utility of this lottery is $p$.

The *standard gamble query* (SGQ) for outcome $o_i$ asks the user to state the probability $p$ for which she would be indifferent between $SG(p)$ and outcome $o_i$ [27]. This type of query is extremely informative as it asks the user to assess a precise tradeoff involving $o_i$, and indeed fixes $u(o_i)$ on the normalized scale. However, this makes such queries practically impossible to answer with confidence. More cognitively plausible are *bound queries*. The bound query $B(o_i, p)$ asks the user whether she would prefer $SG(p)$ to $o_i$. A positive response places an upper bound of $p$ on $u(o_i)$, while a negative response places a similar lower bound. Their yes/no nature makes bound queries easier for people to answer. In principle, bound queries can be used to incrementally elicit utility functions to any required degree of precision by incrementally refining the bounds on utility outcomes, at each stage giving rise to a more refined set of *feasible* utility functions (those consistent with the bounds). In practice, as the feasible regions for each parameter become small, the queries will generally become harder to answer with confidence.

In practice, a lottery is presented as textual (or verbal) description of two outcomes and their probabilities, possibly accompanied by visual aids representing the outcomes. We refer to this delivery of bound query as a *conceptual query*, since people are asked to think about the two alternatives before making a decision. During conceptual elicitation in our domain, each query involves asking the user to think about the two options by imagining the use of two separate systems to complete the highlighting task. $SG(p)$ corresponds to using an adaptive system repeatedly, with percentage $p$ of the instances involving the best interface $o^\top$ and the rest $o_\bot$. Option $o_i$ corresponds to using a static system repeatedly.

Though theoretically appealing, people often have difficulty assessing the probability parameter in SGQs [27]. Similarly, bound queries with precise probabilities can be hard to conceptualize in this domain, with people having difficulty comparing interface "lotteries" with a fixed outcome. Finally, a user's assessment of what they like in hypothetical settings can differ systematically from what they actually like [25; 23; 24; 18]. For these reasons, we investigate alternative elicitation mechanisms that overcome these difficulties.

Preference elicitation in interface customization has typically adopted a qualitative approach to assessing user preferences that learns preference rankings without learning the strength of those preferences [29; 31; 11] (although Horvitz et al. [17] uses "willingness to pay" to quantify the cost of interruption). However, given the stochastic nature of goal estimation, we require estimates of utility as discussed above. To do this within our POMDP, we require some means of associating observed user behavior with utility functions. While direct online elicitation is not (completely) feasible, offline elicitation can be used to develop the required models. For instance, utility functions elicited offline can be clustered into a small set of *user types* [7], which the POMDP assesses online. Online assessment of user types may be done passively [19] or explicitly through active elicitation [4]. In this context, we propose to use an experiential elicitation procedure to carry out offline elicitation experiments with real users, which we describe in the next section.

## 4 Experiential Elicitation

To facilitate interpretation of bound queries, we develop an *experiential* version of bound queries which allows the user to "experience" both query options (including the stochastic one) before stating a preference. The user is asked to complete $k$ simple tasks using the system in each of two ways: the $SG(p)$ option shows $o^\top$ in fraction $p$ of the $k$ tasks and $o_\bot$ in the remaining tasks (in random order). In this way, the user "experiences" a stochastic mixture of interfaces. The deterministic option also requires $k$ task completions, but all using the same interface $o_i$. After each option, the user is asked to reflect on what she liked and disliked about the experience — which could be a function of the effort required by the tasks, toolbar "processing" cost, the satisfaction in having toolbar help available, or the ease of interaction — and to indicate her preference. The response provides a bound at $p$ for $o_i$.

### 4.1 Experiential Elicitation for Interfaces

We elicit $U(N, L, Q)$ from users using the interface shown in Figure 1. We use $k = 10$ tasks for each alternative in the query, so each query involves the user completing 20 highlighting tasks. To reduce the cognitive burden and experiment time, we elicit only $U(N, L, Q)$ for the values $Q \in \{0, 2, 4\}$, $L \in \{1, 5, 10\}$, and $N \in \{0, 1\}$. This discretization yields a range of 18 outcomes. Similarly, we discretize query probabilities: $[0, 0.1, 0.2, ..., 1.0]$. We define $o^\top = n0, l1, q4$, since the user is at a low level of neediness and receives the best help possible, and

$o_\perp = n1, l10, q0$, since the user is at a high level of neediness and receives the worst help possible.

When the system suggests a toolbar, the user can select one icon or ignore it altogether. If the wrong font style is chosen, or the suggestion is ignored, the user must carry out the highlighting task manually. Each task requires the user to carry out several separate actions, and applying an incorrect style requires additional fixes.

### 4.2 Conceptual vs. Experiential Elicitation

We compared experiential and conceptual queries experimentally to investigate their impact on elicitation based on several criteria: the efficiency (duration) of the procedure; the cognitive demand imposed on users; the interpretability or understandability of queries by the user; and the quality of the elicited responses. In particular, we compare the elicited utilities quantitatively under these two conditions, and test whether mean utilities elicited under conceptual elicitation are the same as those under experiential elicitation (our null hypothesis, $H_0$). We also examine the general structure of $U(N, L, Q)$ to see if users perceive value in such simple help, and if any trends across the user population exist in the underlying preferences.

In addition to verbal descriptions, we used screenshots to represent each outcome in the Conceptual condition. Thirteen people participated in the Conceptual condition and 8 people in the Experiential condition.

To control for the user's highlighting goal, we define a *target font style* in each task. As illustrated in Figure 1, each PowerPoint slide has two sentences with the same words, where the top sentence indicates the phrase highlighted with the target font style. The user must match the first sentence by changing the font style of the appropriate words in the second. The vocabulary of target font styles is defined by 7 features—5 of which are binary (bold, underline, italics, shadow, size increment) and 2 of which are multi-valued (8 colors and 10 font families). All target font styles in the experiment have complexity 4. Therefore, if the system's suggestion is perfect, it can save the user from manually executing (sequences of) 4 separate events.

Recall that neediness is a hidden user variable, and how much help a user needs is a function of how difficult the user's goal is to accomplish. In the experiment, we simulate two neediness settings by controlling the task environment, and thus, making the user goal more difficult to achieve. To simulate a needy user ($n1$), we make the task more difficult by restricting the set of colors to 7 shades of red and the set of font families to 4 similar fonts. In this way, the system's feature vocabulary, the target font styles in the highlighting tasks, and the icons in the toolbar are restricted to similar colors and fonts. The interface in Figure 1 shows an example of this neediness setting. The full default feature vo-cabulary defines the interface environment for a user who is not needy ($n0$).

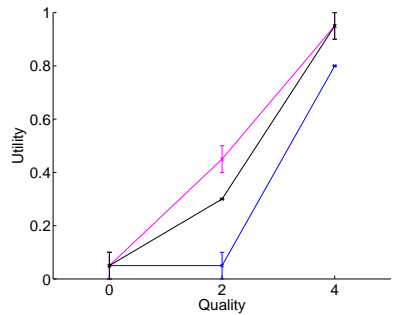

Figure 2: Partial utility functions for $n0, l10$ as a function of $Q$ elicited from 3 users in the Experiential condition. Lines are drawn at the midpoints of the resulting bounds.

During the elicitation, we posed bound queries and incrementally refined the bounds by choosing $p$ to be the midpoint of the set of feasible utility functions until all the outcomes have feasible regions with range $\leq 0.1$. Figure 2 illustrates sample results from a specific elicitation run, showing a partial utility function (for fixed values of $N$ and $L$) for three users. The elicited bounds are drawn as error bars (note that discretization prevents us from pinning down the utility function precisely). The "blue" user (lower line), for instance, has $0 \leq u(n0, l10, q0) \leq 0.1$. The bound at $q4$ for the blue user is tight because she is indifferent between $SG(.8)$ and $o_i = n0, l10, q4$.

**Methodological Comparison** On average, each experiment took 30 minutes in the Conceptual condition and 2 hours (divided into two sessions) in the Experiential condition. Experiments in the Experiential condition are much longer because experiential queries require users to carry out tasks, while the conceptual queries only require users to think about scenarios.[3]

Since experiential queries require a series of task completions, users became tired early on and found it necessary to take breaks in order to not be confused with the various options and associated experiences. Users in the Conceptual condition did not seem tired during the procedure, but they were at times inconsistent with previous responses.

Although experiential queries took longer, they provided hands-on experience and therefore required little verbal explanation. In contrast, conceptual queries were often diffi-cult to explain, because they require users to first understand the aspects of the interface (e.g., amount of effort needed in manually completing the tasks, controlled quality of help in the suggestions), then compare the costs and benefits of a mixture of interfaces with a definitive interface, and finally, imagine using the respective interfaces in a repeated scenario.

---

[3]Our aim is to engage in *complete* utility elicitation (to the prescribed accuracy) to develop models of (classes of) user preferences that can be used in online assessment (e.g., within a POMDP). For this reason, we do not consider means to "intelligently" assess only the *relevant* preferences, using, say, value of information with respect to a specific task, something which is vital in online assessment. Of course, if general domain constraints are known to render various outcomes impossible, we could prune the elicitation task somewhat.

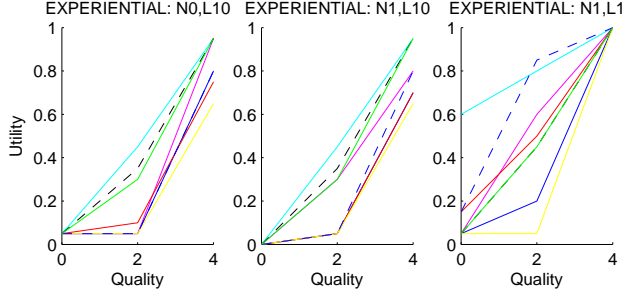

Figure 3: Partial utility functions for $n0, l10$, $n1, l10$, and $n1, l1$ as a function of $Q$ elicited from all 8 users in the Experiential condition. Each line connects the midpoints of the outcome's elicited bounds for each user.

**Structural Comparison** Independently of the elicitation method, we are interested in the perceived value of this adaptive form of customization. The utility functions across the 21 users varied widely—some are convex, some are concave, some are linear, some are "flat" when quality is not perfect, and some are "flat" when length is not one. Some examples (using midpoints of feasible regions) are shown in Figure 3. Clearly, user preferences vary widely, even for such simple highlighting help with three customization attributes.

In general, the utility functions are monotonically non-decreasing in $Q$ when $N$ and $L$ are fixed, and monotonically non-increasing in $L$ when $N$ and $Q$ are fixed. In particular, when help quality is high ($q4$), utility decreases slightly as $L$ increases. This is expected as users perceive higher processing costs with more icons. We also see that *partial help* ($q2$) with $L$ at $l1$ is qualitatively different than $l5$ or $l10$, because more icons decrease the chance of the user identifying the single icon that provides partial help. When help quality is low ($q0$), some users prefer to see one bad suggestion ($l1$) than many bad suggestions ($l10$). There are no general trends in utility given neediness; some users showed clear differences between the needy ($n1$) and not needy ($n0$) scenarios, while others viewed them the same. From this, we see that users perceive value in automated help, even in simple tasks such as highlighting. Of course, more data is needed to draw definitive conclusions about possible parametric forms for utilities that could further simplify online assessment.

**Quantitative Comparison** Using Hotelling's $T^2$ statis-

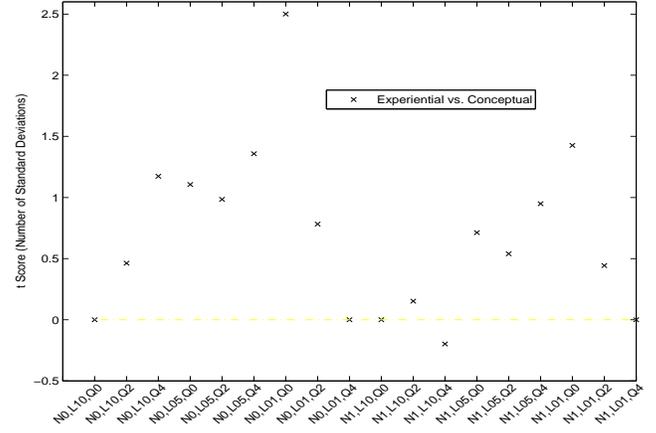

Figure 4: Resulting $t$ scores for each elicited outcome.

tic,[4] we found that the mean utilities in the two conditions are significantly different, $p < 0.01$. Thus, we reject $H_0$. To provide a more detailed comparison, we carried out a two-tailed $t$-test with independent means for each outcome. The resulting $t$ scores[5] are plotted in Figure 4. Only one outcome, $n0, l1, q0$, is *individually* significantly different between the two conditions (with 19 degrees of freedom, a $t$ score of at least $\pm 2.093$ is needed for significance at $p < .05$).

More interestingly, Figure 4 shows that the mean $t$ score tends to be higher for most outcomes in the Experiential condition, including outcomes that provide partial quality help ($q2$) and incorrect suggestions ($l5, q0$ and $l1, q0$). This indicates that users in the Experiential condition perceive greater value in adaptive help than users in the Conceptual condition. The Experiential condition requires users to carry out 20 tasks for each query, while the Conceptual condition only asked users to "think about" the tasks. With conceptual queries, we believe that participants are less likely to truly perceive the value of automated help in repeated scenarios, and thus, underestimate the utility of these outcomes. We believe that experientially assessed utilities more accurately reflect the users' true preferences; however, our experimental set up does not allow us to draw such a definitive conclusion.

## 5 Ways to Improve the Experiential Elicitation Procedure

Although it seems that experiential queries enable users to report more realistic preferences, the procedure is time consuming (even with simple utility functions over 18 out-comes). Though our intent is to examine methods for offline elicitation to support models for online adaptation—thus we do not face the demands on online customization here—even for offline model development this procedure may be too demanding. We develop two more efficient elicitation procedures, based on the findings in Section 4.2. Following the same experimental set up, we introduce two procedures, *primed* and *primed+*, that attempt to elicit experienced utility more effectively. The Primed condition uses a training session to familiarize users with the interface and the attributes $N, L, Q$; but the elicitation procedure itself still relies on conceptual queries only. The Primed+ condition uses this training session plus 5 experiential queries at the start of the elicitation. The remaining elicitation is done using conceptual queries only.

---

[4]The $T^2$ distribution is a multivariate analog of the Student's $t$-distribution. We used the Moore-Penrose pseudoinverse in computing $T^2$.

[5]A $t$ score is a measure of how far apart the two sample means are on a distribution of differences between means.

Similar to the previous experiment, we want to test whether the mean utilities elicited under the Conceptual condition are the same as those elicited under the Primed and Primed+ conditions (our null hypothesis $H_0$). A total of 9 and 8 people participated in the Primed and Primed+ conditions respectively.

**Methodological Comparison** Both procedures were easier to administer than the experiential and conceptual ones. First, the familiarity acquired in the training session reduced the need to explain the interface to the users. On average, each experiment took 30 minutes in the Primed condition and 60 minutes in the Primed+ condition. Neither conditions seemed tiring for users. Second, users in both conditions found the queries easier to understand than users in the Conceptual condition. Finally, users in the Primed+ condition were able to use their experiential query responses as a reference for future responses. These initial experiential queries provided users with a quick feeling for the sequential costs and benefits of using the toolbar.

**Quantitative Comparison** We conducted a pairwise analysis between the mean utilities in the Conceptual and Primed conditions, and between those in the Conceptual and Primed+ conditions. Using Hotelling's $T^2$ statistic, we found that the mean utilities between Primed/Primed+ and Conceptual conditions are significantly different ($p < 0.01$ and $p < 0.05$ respectively). Thus, we reject $H_0$ in both instances. The results of a pairwise analysis using a two-tailed $t$-test with independent means for each outcome are shown in Figure 5. None of the outcomes individually are significantly different between the new conditions and the Conceptual condition.

From Figure 5, we see that the primed means are generally lower ($t$ score less than 0) than conceptual. In fact, the $t$ scores for the Primed condition (vs. Conceptual) are always lower than the $t$ scores for Experiential (vs. Conceptual). One explanation for this is the fact that the training session in the Primed setting gave users a quick estimate of the costs of searching through and evaluating suggestions in the toolbar—a cost that would otherwise be "unknown" in the Conceptual condition—but that the short experience was insufficient to provide the user with a sense of the *benefits* of help. In both the Primed and Conceptual conditions, the cognitive demand involved in repeated highlighting tasks and the value of help is consistently underestimated. In contrast, the Primed+ condition and its five experiential queries provide a sense of long-term benefits of using the toolbar. Indeed, in Figure 5, the general pattern indicates that Primed+ approaches Experiential. These results support our initial hypothesis that experiential queries enable the user to perceive the full value of adaptive help under realistic circumstances.

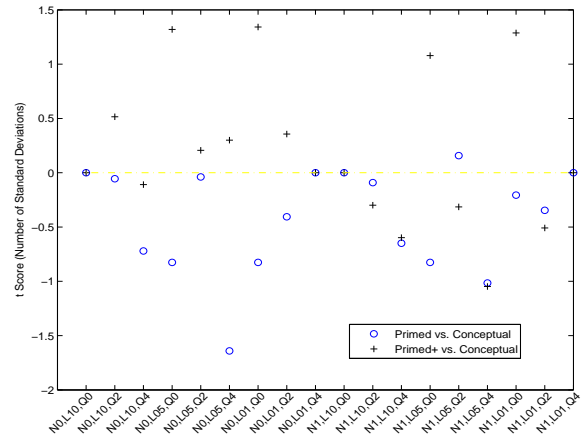

Figure 5: Resulting $t$ scores for each elicited outcome.

## 6 Conclusions and Future Work

Traditional approaches to utility elicitation do not seem to be effective in assessing the preferences of users for interface customization and adaptive help design, largely due to the lack of experiential assessment. We have proposed a new experiential elicitation procedure that is well-suited to this and related tasks. Our results show that experiential elicitation has several benefits, including: ease of administration from the researcher's perspective; understanding of outcomes from the user's perspective; and helping users appreciate the sequential nature of interaction, often overlooked in traditional, "conceptual" elicitation. We also developed the primed+ procedure to speed up experiential elicitation while maintaining most of its benefits.

In this paper, we focused on eliciting a utility function with three attributes that model savings and processing cost in the context of interface customization. In general, a user's utility function may involve other attributes, depending on the possible customization actions. For example, an adaptive system that hides unused functions causes *disruption* to the user's mental model of the application, but reduces interface *bloat*. By adopting the methodology illustrated in this work, analogous experiments can be devised to ex-perientially elicit user preferences over these attributes for interface customization.

Future plans include gathering more data to potentially learn a parametric form for the utility function $U(N, L, Q)$. Intuitively, our results suggest a quadratic functional form may explain most preferences, but more data is needed to draw definitive conclusions. We are also interested in examining the extent to which a utility function of this form can be applied more generally to different customization and help tasks. This would allow for the learning of individual user utility models that apply to multiple tasks and even multiple applications. Finally, we are interested in the extent to which lessons in offline elicitation influence the development of online active elicitation and utility assessment strategies, especially the development of behavioral and query response models (e.g., for a POMDP).

Further development of experiential elicitation will require better understanding of which aspects of the outcomes make them experientially different (either better or worse) from a user's conceptual prediction. For example, Figure 4 indicates that, on average, the outcome $n1, l10, q4$ is actually not as good as people think after experiencing it. One explanation is that users did not expect much difficulty in searching for a matching icon ($q4$) when neediness is high. When compared to the mean utility of $n0, l10, q4$, we see that users underestimate the value of help when they are needy. We believe more interesting patterns will unfold in richer domains (i.e., with more attributes and outcomes) and with more data.